\pdfoutput=1

\documentclass[tikz,11pt]{article}

\usepackage{natbib}
\usepackage[final]{acl}

\usepackage{times}
\usepackage{latexsym}
\usepackage{float}
\usepackage{amsmath}

\usepackage[T1]{fontenc}
\usepackage[utf8]{inputenc}
\usepackage{microtype}
\usepackage{inconsolata}

\usepackage{graphicx}
\title{
    POMO+: Leveraging starting nodes in POMO for solving \\
    Capacitated Vehicle Routing Problem
}

\author{
        Szymon Jakubicz \\ 
        University of Warsaw, Poland \\
        \texttt{s.jakubicz@student.uw.edu.pl} 
        \And
        Karol Kuźniak \\ 
        University of Warsaw, Poland \\
        \texttt{km.kuzniak@student.uw.edu.pl} 
        \AND
        Jan Wawszczak \\ 
        University of Warsaw, Poland \\
        \texttt{janekw23@gmail.com}
        \And
        Paweł Gora \\
        Fundacja Quantum AI \\
        \texttt{pawel.gora@qaif.org} \\
        \AND
}

\begin{document}
\maketitle

\begin{abstract}
In recent years, reinforcement learning (RL) methods have emerged as a promising approach for solving combinatorial problems. 
Among RL-based models, POMO has demonstrated strong performance on a variety of tasks, including variants of the Vehicle Routing Problem (VRP). 
However, there is room for improvement for these tasks. In this work, we improved POMO, creating a method (\textbf{POMO+}) that leverages the initial nodes to find a solution in a more informed way. 
We ran experiments on our new model and observed that our solution converges faster and achieves better results. 
We validated our models on the CVRPLIB dataset and noticed improvements in problem instances with up to 100 customers. 
We hope that our research in this project can lead to further advancements in the field.
\end{abstract}

\section{Introduction}

The Vehicle Routing Problem (VRP) generalizes the Traveling Salesman Problem (TSP) by seeking the minimum-cost set of routes for a fleet of vehicles that start and end at a depot while satisfying customer demands \citep{Dantzig1959}.
This problem can be applied in real‐world logistics, where even small improvements translate into large savings—industrial studies report 5–30\% lower transport costs after route optimization \citep{Hasle2007}.  
Because VRP is NP-hard, exact branch-and-cut or column-generation algorithms scale only to tens of customers, so research has long relied on heuristics, beginning with the classical savings algorithm of \citet{ClarkeWright1964} and ending on genetic algorithms or large-neighbourhood search \citep{Toth2002}.

Operational constraints give rise to a rich family of problems. A recent survey lists more than fifty VRP variants \citep{Vidal2020Concise}.  
The Capacitated VRP (CVRP)—in which each vehicle has a limited capacity — is both one of the most studied and the most often deployed variant in practice, so we put our focus on it.

Over the past years, Machine-Learning (ML) and, in particular, Reinforcement Learning (RL) have emerged as a serious alternative to hand-crafted heuristics.  
Attention-based policies trained with RL already match specialized solvers on 100-node instances while offering orders-of-magnitude faster inference.
Models based on the Attention Model \citep{Kool2019Attention} and POMO \citep{Kwon2020POMO} now set the bar for neural VRP heuristics.

Guided by these advances, our work reviews state-of-the-art RL solvers, selects the most promising one to improve (POMO), and enhances it with a lightweight auxiliary agent that learns to choose the best starting nodes, which is an open suggestion from the authors of the original model.
We called our method POMO+, implying, that this approach, does not necesserily change the main idea of POMO, but strenghten this.
The resulting method delivers better solutions on CVRP benchmarks while maintaining POMO’s training efficiency.

\section{Related Work}
Early neural methods, such as Pointer-Networks \citet{vinyals2017pointernetworks}, framed routing as a sequence-generation task.
These models later stepped aside, for models that integrate attention mechanisms to solve routing problems more efficiently.
The \emph{Attention Model} (AM) of \citet{Kool2019Attention} replaced Pointer-Networks with a Transformer encoder–decoder and, trained with the REINFORCE algorithm, delivered near-optimal solutions for TSP and the first competitive learned heuristics for CVRP.

On the other hand, \citet{Chen2019NeuRewriter} showed that construction heuristics can be refined after decoding.  
Their \emph{NeuRewriter} agent uses actor-critic learning to choose \emph{where} and \emph{how} to locally rewrite a partial solution, consistently outperforming classical heuristics and earlier neural baselines on VRP.

Furthermore, there exists models that approximates solutions from heuristics for these problems.
\citet{Kool2022DPDP} proposed \emph{Deep Policy Dynamic Programming} (DPDP), which uses a neural edge-selection policy to prune the DP (Dynamic Programming) state space, scaling dynamic programming to 100-node instances while remaining competitive with LKH \cite{helsgaun2000effective}. 

One of the state-of-the-art pure-RL baselines in this field is \emph{Policy Optimisation with Multiple Optima} (POMO) by \citet{Kwon2020POMO}.  
POMO exploits permutation symmetry by launching one rollout from every node and averaging their costs as a low-variance baseline using the REINFORCE algorithm, which speeds up training and improves robustness.
However, the authors of the model, notice that for VRP problems, there is room for improvement during training by choosing from which nodes the rollout will be launched.
In the next section, we will focus more on explaining this problem, and our approach to solving it.

Worth mentioning is also other POMO-based models. For example, \citet{Gao2024ELG} combined a global construction policy with a transferable local policy in an \emph{Ensemble of Local and Global} (ELG) framework, achieving strong cross-distribution results on TSPLIB \citet{Reinelt1991} and CVRPLIB \citet{UCHOA2017845}. 
This approach, with adding additional small improvement inspired us to add an auxiliary agent in POMO.

Complementary to algorithmic advances, the \emph{RL4CO} \citep{Berto2025RL4CO} unifies 23 solvers—including AM, POMO and ELG—which we adopt as baselines. This work unifies recent developments in this field and provides a simple framework for implementing and testing improvements of these models. We used this framework to adopt our solution.

\section{Approach}

\subsection{Problem description}

The Capacitated Vehicle Routing Problem (CVRP) is a variant of the Vehicle Routing Problem in which, given a set of clients with known demands and a maximum vehicle capacity, the goal is to determine the optimal number of vehicles and their routes so that each client is served exactly once, vehicle capacities are not exceeded, and the total travel distance is minimized.

Similarly to the POMO paper, let $N$ be the number of customers (nodes), denoted as $\{ v_1, \dots, v_N \}$, with $v_0$ representing the depot and $C$ the maximum vehicle capacity.  
A \emph{trajectory} $\tau$ is a sequence representing a valid traverse of clients in terms of the CVRP:
\[
\tau = (a_1, a_2, \dots, a_M),
\]
where each $a_i$ denotes visiting node $v_j$.  
The depot $v_0$ may appear multiple times in the trajectory, indicating the end of one vehicle’s route and the start of another.

A trajectory is created autoregressively by a model with parameters $\theta$:
\[
\pi_t = 
\begin{cases}
p_\theta(a_t \mid s) & \text{for } t = 1, \\
p_\theta(a_t \mid s, a_{1:t-1}) & \text{for } t \in \{2, 3, \ldots, M\},
\end{cases}
\]
where $s$ represents the state of problem instance.

\subsection{POMO approach}

POMO is essentially an Attention Model that generates multiple trajectories simultaneously, rather than a single trajectory, as commonly done before.
Initially, POMO selects \(N\) different starting points $\{a_1^1, a_1^2, \dots, a_1^N\}$ which correspond to all clients. The network then generates \(N\) trajectories using a Monte Carlo method within a reinforcement learning framework. The \(i\)-th trajectory is denoted as
\[
\tau^i = (a_1^i, a_2^i, \dots, a_M^i),
\]
where \(M\) is the number of visited nodes (including visits to the depot).

Training in POMO is based on the REINFORCE algorithm.
Once the trajectories $\{\tau^1, \dots, \tau^N\}$ are obtained, we calculate the reward $R(\tau^i)$ for each trajectory $\tau^i$.
The reward is defined as the negative total travel distance of the trajectory.
By maximizing this reward, the model effectively minimizes the total distance, thus approaching the optimal solution.
Let $L$ represent the expected reward.
Then its gradient with respect to the model parameters \(\theta\) is approximated by
\[
\nabla_{\theta} L(\theta) \approx \frac{1}{N} \sum_{i=1}^N \left[ \left(R(\tau^i) - b^i(s)\right) \nabla_{\theta} \log p_{\theta}(\tau^i \mid s) \right],
\]
where $
p_{\theta}(\tau^i \mid s) = \prod_{t=1}^M p_{\theta}(a_t^i \mid s, a_{1:t-1}^i)$,
and \(b^i(s)\) is a shared baseline for all trajectories, defined as
\[
b^i(s) = b_{\text{shared}}(s) = \frac{1}{N} \sum_{j=1}^N R(\tau^j).
\]

To conclude, POMO employs the Attention Model as its policy network, which consists of an encoder and a decoder.
The encoder produces contextualized embeddings for each node and the decoder generates the solution autoregressively using these embeddings with the attention mechanism.
This is a brief overview of the model, but it is enough to give you the outline of our improvements. 
For implementation details, we refer the reader to the original POMO and Attention Model papers.

\subsection{Our approach}



In the original POMO paper, the authors point out that in the CVRP — unlike in the TSP — selecting good initial starting points for creating trajectories is crucial.
In the TSP, there is no difference in which node is visited first, because the optimal solution is a cycle.
On the other hand, in VRP variants, there are multiple vehicles, and each vehicle starts at the depot.
Thus, if, for example, the starting node is \(v_1\), then one of the vehicles will first visit \(v_1\), which imposes an additional constraint.
In POMO, the model creates \(N\) trajectories, each trajectory beginning at one of the \(N\) clients.
For each trajectory, the starting node becomes the first customer visited by the first vehicle.
When a vehicle returns to the depot — either due to capacity constraints or the model’s decision — that vehicle’s route ends.
If some nodes are still not visited, the model selects the next starting point of a route for a new vehicle.
Because the initial customer directly affects the whole trajectory, starting from certain nodes can lead to exploring inefficient routes.
This is an exploration problem: some starting points could be much worse than others.
To address this problem, we added a lightweight agent to choose starting nodes for VRP variants.
Our auxiliary agent, unlike, e.g., ELG, is trained alongside POMO and uses the full context of each problem instance using the AM encoder.

\subsubsection{Training}

For set of trajectories $\{\tau^1, \tau^2, \dots, \tau^N\}$, the agent with parameters $\omega$ is trained using the REINFORCE algorithm:
\[
    \nabla_{\omega} L(\omega) \approx \frac{1}{N} \sum_{i=1}^{N} \left[ \left( R(\tau^i) - b(s) \right) \nabla_{\omega} \log p_{\omega}(a_1^i | s) \right]
\]
with the same notation as introduced earlier.   

Note that $\omega$ refers only to the parameters of the auxiliary model; this does not affect the original encoder.

Our model is trained alongside POMO. To stabilize training, we accumulate gradients for the auxiliary model over $n$ steps and apply an update after each accumulation. For our experiments, we set $n = 100$ for all models.

\subsubsection{Architecture}
\begin{figure}[H]
  \centering
  \includegraphics[width=0.2\textwidth]{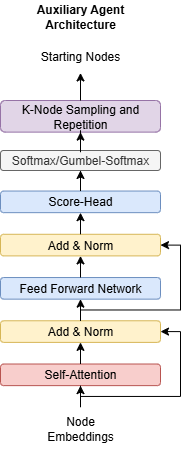}
  \caption{Architecture of auxiliary agent to select best starting nodes for POMO. The embedded nodes are given by the encoder, the same values are passed to POMO.}
  \label{fig:architecture}
\end{figure}
\begin{figure}[H]
  \centering
  \includegraphics[width=0.5\textwidth]{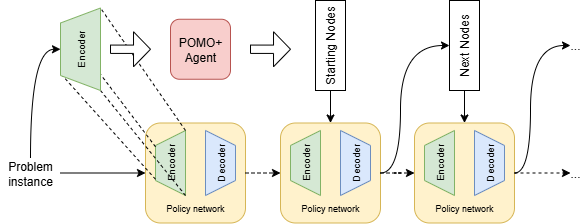}
  \caption{Visualization of our agent within the overall POMO architecture. It operates on embeddings from policy network encoder.}
  \label{fig:arch_whole}
\end{figure}

The architecture of our agent (see Figure~\ref{fig:architecture}) is simple and leverages the hidden representations produced by the Attention Model Encoder of the POMO architecture (see Figure ~\ref{fig:arch_whole}).
This approach, instead of creating a separate attention encoder, is much more efficient than training two distinct encoders and does not significantly increase the training time.

Our agent takes as input the hidden representations of all nodes (including the depot) produced by the encoder.  
During the backward pass, the encoder parameters remain fixed, and only the agent's parameters are updated.  
The representations are passed through our lightweight agent, which consists of a Multi-Head Attention layer and a feedforward network, with residual connections and normalization after each block.  
The output is then processed by a scoring head composed of linear layers to compute a score for each node.  
An activation function — Softmax during inference or Gumbel–Softmax during training — is applied to obtain a probability distribution over nodes (due to the non-differentiable property of the softmax function, we use Gumbel–Softmax during training and softmax during inference).  

Finally, the agent samples \(K\) starting nodes from this distribution and repeats each of them \(N / K\) times to match the total number of starting nodes with the number of clients (\(N\)). To avoid bias, we always choose \(K\) such that it divides \(N\), ensuring that the \(N / K\) trajectories share the same starting node. Therefore, the number of all trajectories is the same as in POMO, and is equal to the number of clients ($N$). Thus, the set of starting points of trajectories \(\{a_1^1, a_1^2, \dots, a_1^N\}\) contains only \(K\) distinct values.

\subsection{Inference}

Inference is performed by sampling with softmax. We also select the best result from 8 augmented problem instances created using basic transformations such as rotations and reflections, as in the original POMO.

\section{Experimental results with analysis}
\subsection{Experimental setup}

We trained three separate models on CVRP instances with \(N\in\{20,50,100\}\) customers. Our experiments were run on the ICM UW cluster called \textit{Rysy}.

For every \(N\), we generated \(100{,}000\) synthetic instances following the sampling procedure from the original POMO work and kept an additional \(10{,}000\) instances for testing. 
Both our method and the baseline were trained with a batch size of 64.

\subsection{Baseline}
The baseline is the unmodified \textbf{POMO} model, and our enhanced model is referred to as \textbf{POMO+}.
In POMO+, the auxiliary start-node policy selects \(0.2N\) candidates (4, 10, or 20, respectively) and is updated every 100 environmental steps, as mentioned earlier.
Both were implemented in the \textit{rl4co} framework and share all hyperparameters reported in the POMO paper.

\subsection{Evaluation}
We evaluated on two datasets:
\begin{itemize} 
    \item A - the 10{,}000 test instances, where the depot and customer locations are randomly sampled within the square $[0, 1]^2$. Customer demands are drawn uniformly from the range $[0, 10]$. The vehicle capacity $C$ is set to 30, 40, and 50 for instances with $N = 20, 50,$ and $100$ customers, respectively.
    \item B - \textbf{CVRPLIB} benchmark, where problems vary in size from 10 to over 300 customers, and includes instances from all of the most important datasets.
\end{itemize}
 
Performance on \textbf{CVRPLIB} is measured by the average \textbf{gap} — the percentage difference between the cost returned by the model and the best known solution.
As \textbf{CVRPLIB} records the best known solutions for the fixed number of vehicles, while POMO models are free to use any number of delivery trucks, we treat those best known solutions as a reference baseline rather than an absolute optimum.

\subsection{Results}

\subsubsection{Validation training curves}
\begin{figure}[H]
  \centering
  \includegraphics[width=0.48\textwidth]{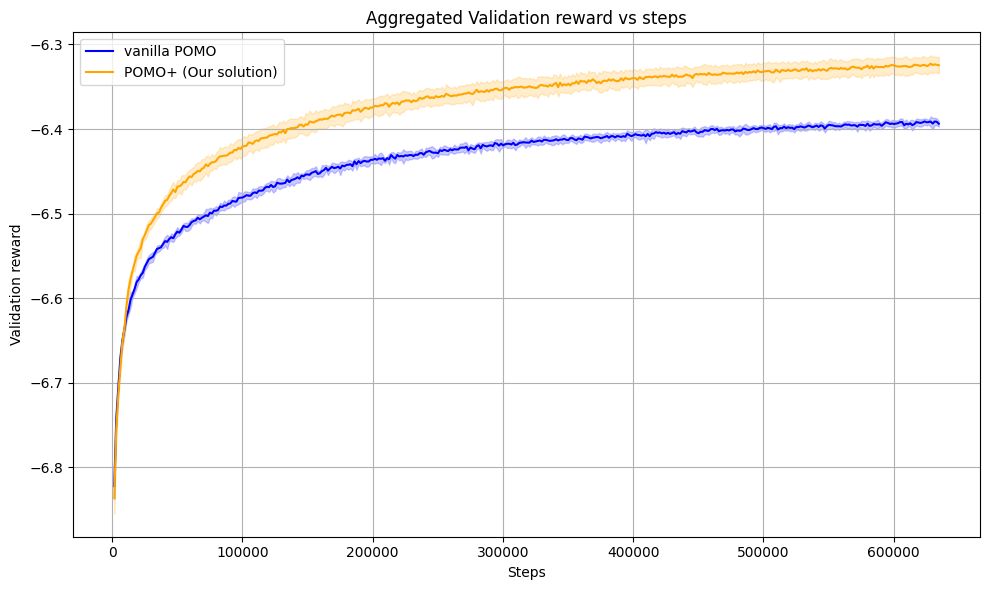}
  \caption{Average validation cost during training on the dataset A, $N{=}20$.}
  \label{fig:train20}
\end{figure}

\begin{figure}[H]
  \centering
  \includegraphics[width=0.48\textwidth]{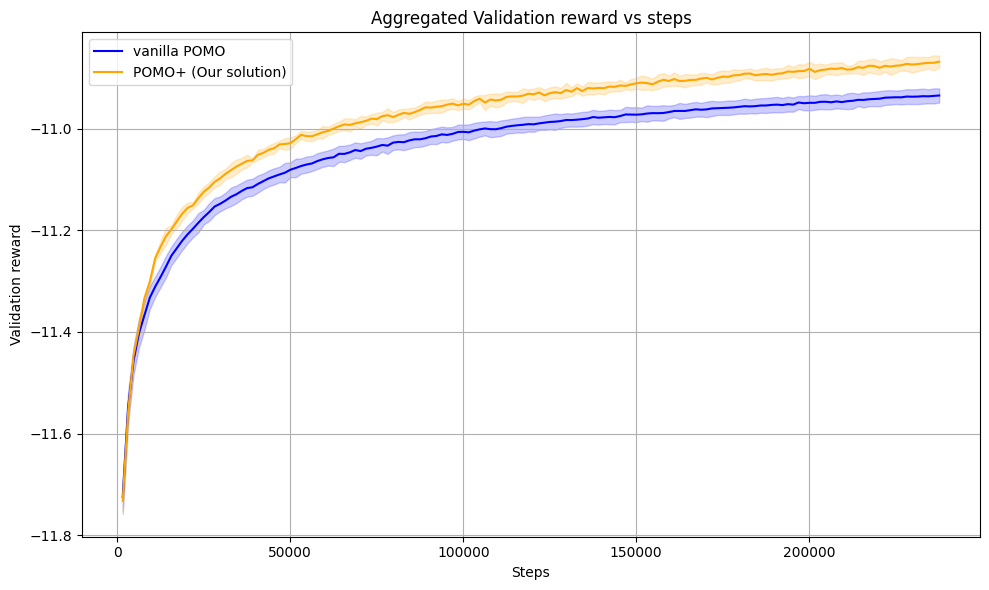}
  \caption{Average validation cost during training on the dataset A, $N{=}50$.}
  \label{fig:train50}
\end{figure}

\begin{figure}[H]
  \centering
  \includegraphics[width=0.48\textwidth]{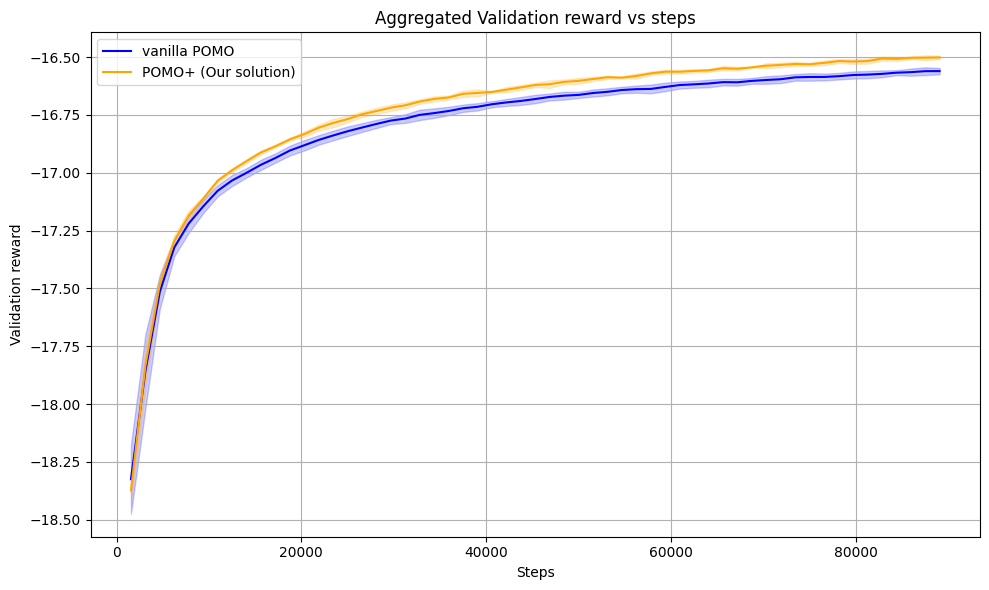}
  \caption{Average validation cost during training on the dataset A, $N{=}100$.}
  \label{fig:train100}
\end{figure}

Figures~\ref{fig:train20}–\ref{fig:train100} plot the validation reward for the dataset A during training, which, for both models, is the negative cost (length) of all trajectories.
This means that higher (less negative) values indicate better solutions.

For all three problem sizes (\(N=20,50,100\)), the POMO+ curve rises more steeply than the vanilla POMO curve and stabilizes at a higher plateau.  
The performance gap is largest for \(N{=}20\) and gradually narrows as \(N\) increases, but POMO+ remains on top throughout.

\subsubsection{Evaluation on \textsc{CVRPLIB}.}
\begin{figure}[H]
  \centering
  \includegraphics[width=0.48\textwidth]{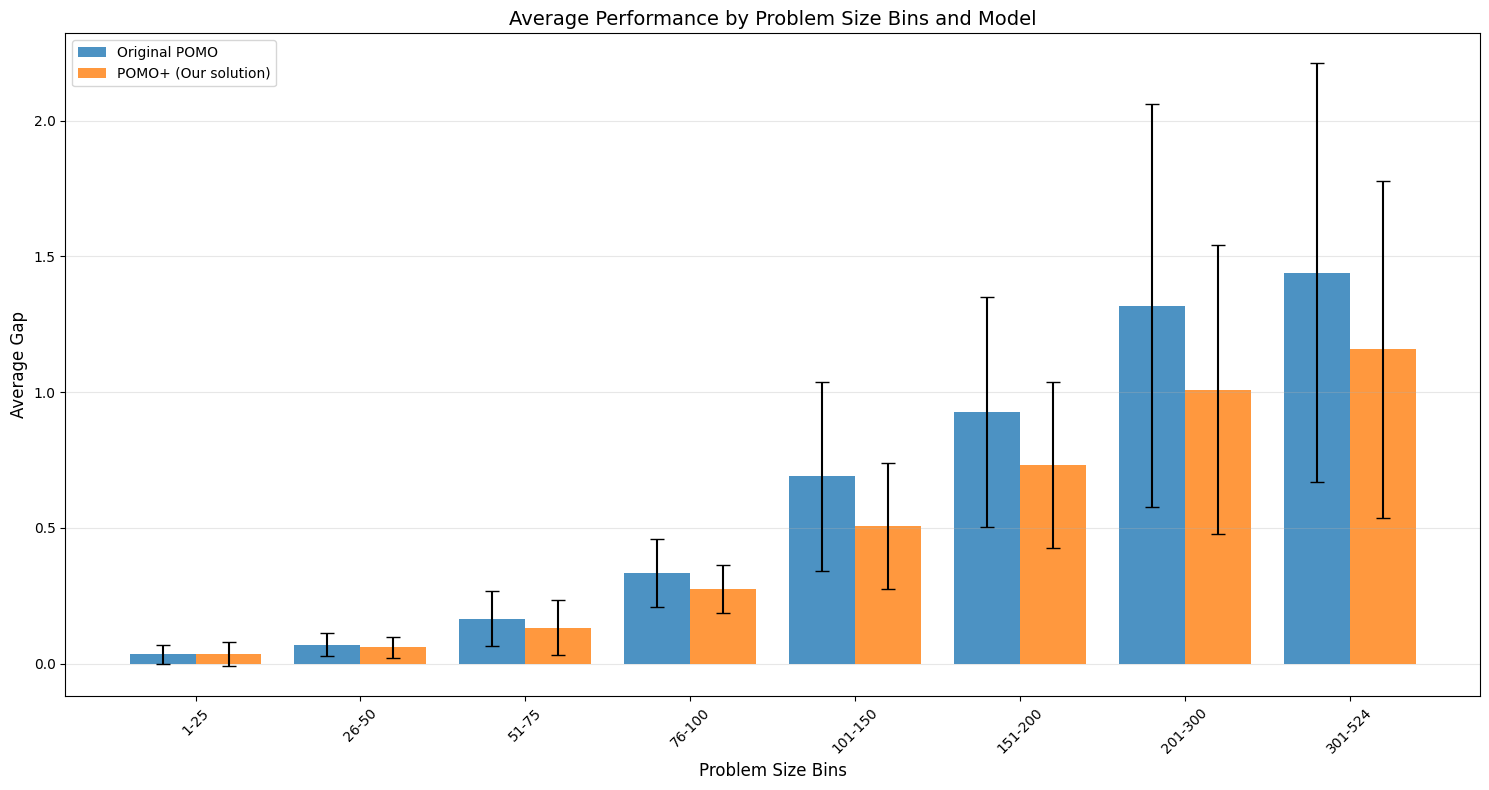}
  \caption{Optimality gap on \textsc{CVRPLIB}, models trained with $N{=}20$.}
  \label{fig:cvrlib20}
\end{figure}

\begin{figure}[H]
  \centering
  \includegraphics[width=0.48\textwidth]{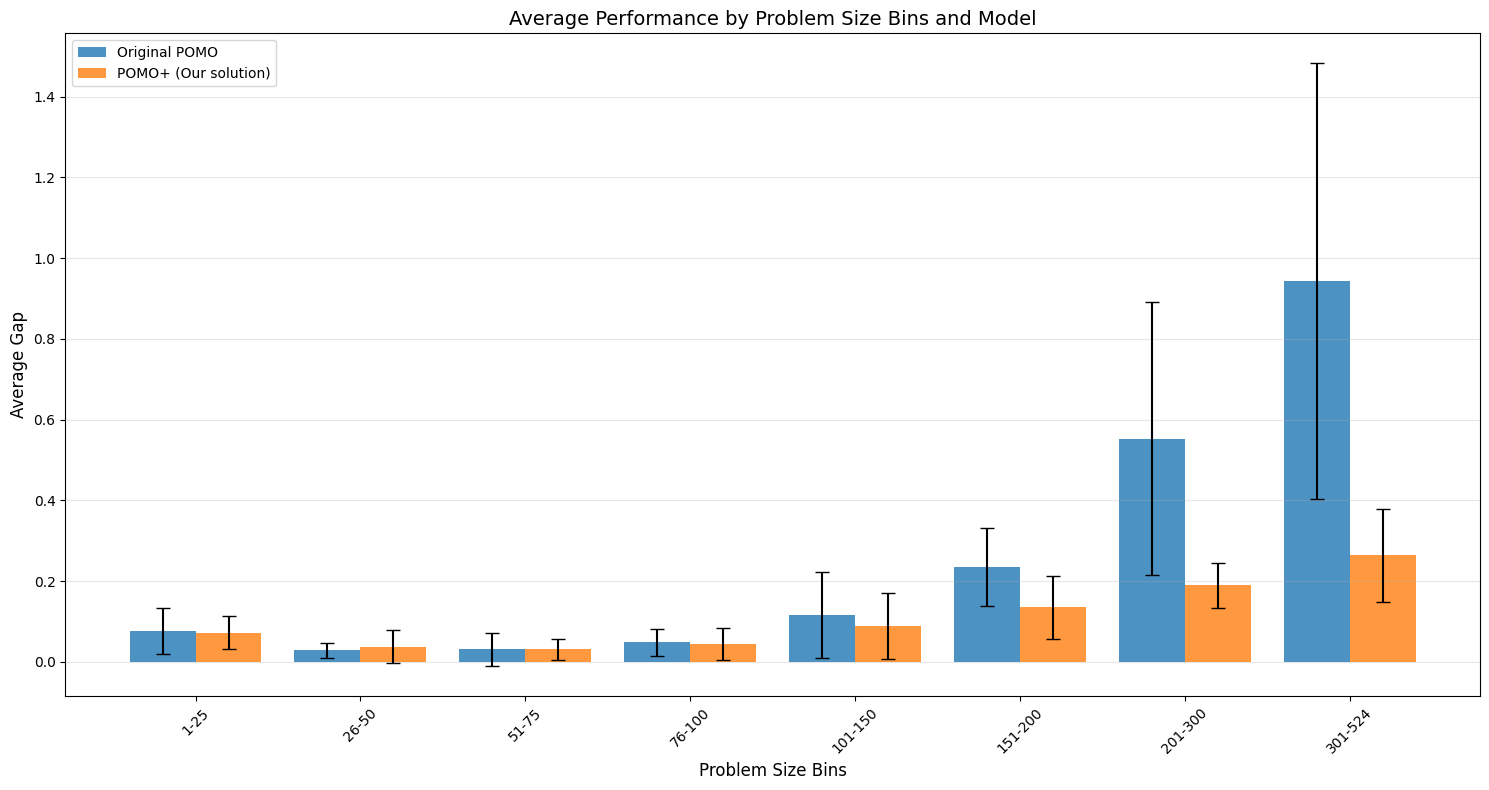}
  \caption{Optimality gap on \textsc{CVRPLIB}, models trained with $N{=}50$.}
  \label{fig:cvrlib50}
\end{figure}

\begin{figure}[H]
  \centering
  \includegraphics[width=0.48\textwidth]{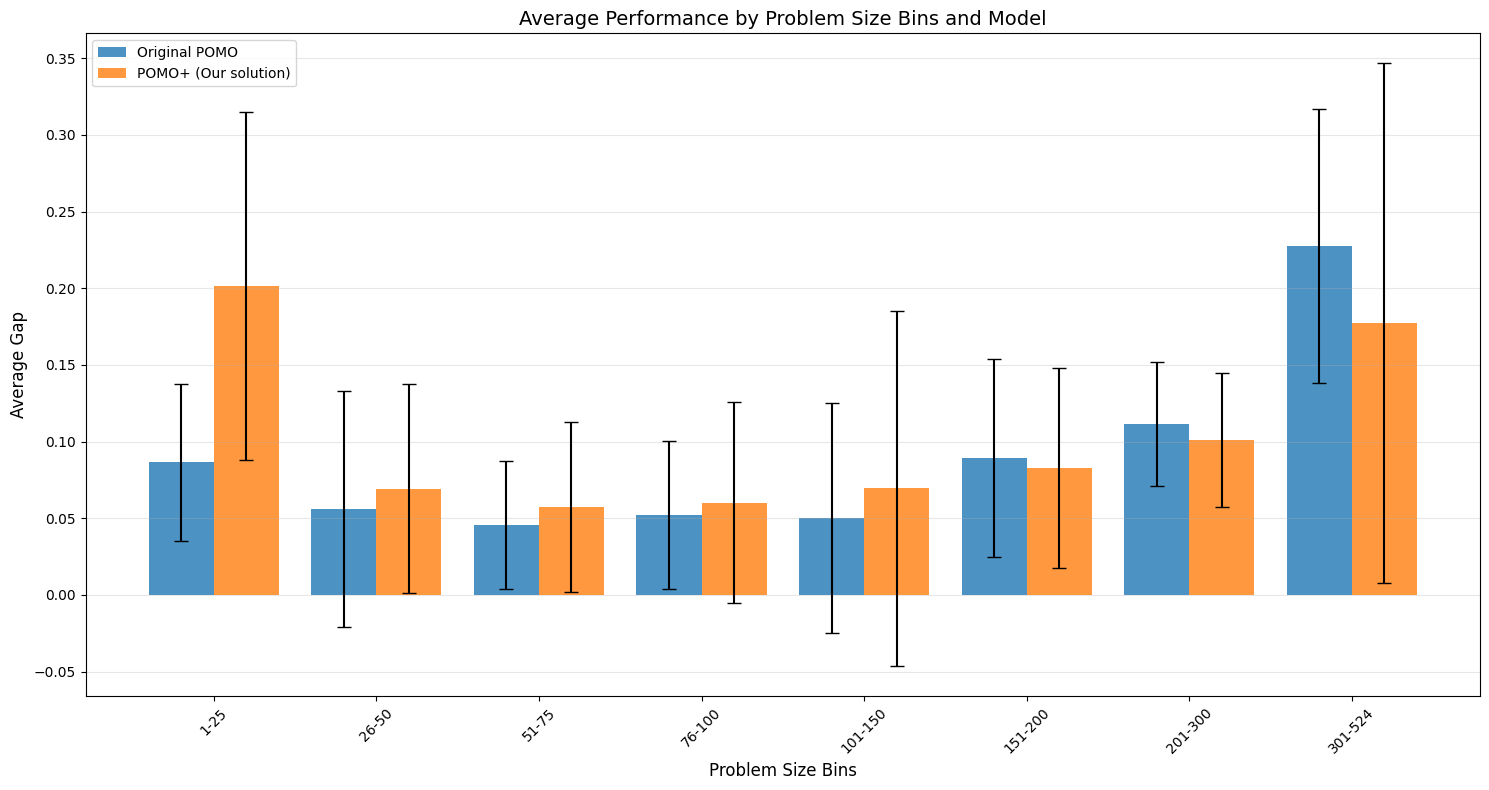}
  \caption{Optimality gap on \textsc{CVRPLIB}, models trained with $N{=}100$.}
  \label{fig:cvrlib100}
\end{figure}

Figures~\ref{fig:cvrlib20}–\ref{fig:cvrlib100} report the average optimality gap on the \textsc{CVRPLIB} benchmark, grouped by instance‐size ranges, for $3$ models trained on problem sizes $N=20,50,100$, respectively.
\clearpage
\begin{itemize}\setlength\itemsep{2pt}
    \item \(N = 20\) \\
          POMO+ achieves a lower gap than POMO for every size range; however, optimality gaps are still significant for bigger instances.
    \item \(N = 50\) \\
          The advantage of POMO+ widens, especially on the larger instances. Gaps are smaller than for \(N=20\)
    \item  \(N = 100\) \\
          Both models are the best among all models.  
          POMO is better for solutions up to 150 customers, however, for larger instances, our solution is better.
\end{itemize}

\subsubsection{Analysis}
The validation reward curves show that our solution is superior compared to the original approach. 
Evaluation on \texttt{CVRPLIB} for $N = 20$ and $N = 50$ customers also shows significant improvement with our model.
However, the results on \texttt{CVRPLIB} for $N = 100$ are ambiguous, making it difficult to determine whether our model scales better.
It is important to note that we have not tested different values of $K$ for these models, and our largest model may require more epochs to train properly.

\section{Conclusions and Future Work}
In our project, we investigated RL methods for solving VRP problems. 
We identified one of the best RL models from recent years, POMO, which still had some room left for improvement, and we developed a new method to enhance this model. 
Our experiments produced promising results for future work and highlighted the need for a deeper understanding of why this method performs well. 
We also observed that our models demonstrate better scalability to large instances compared to the original POMO model. 

Promisingly, our approach offers a lightweight enhancement for models based on POMO. 
Future work could include testing our method on POMO-based models such as OMNI-POMO \citet{gao2024generalizableneuralsolversvehicle} or ELG, as well as combining it with techniques like leveraging leader rewards \citet{wang2024leaderrewardpomobasedneural}. 
Additionally, our study did not include an ablation analysis of the choice of the optimal number of starting nodes, nor did it explore different VRP variants, nor trainings on datasets that vary in size. 
These research directions could lead to significant improvements in RL methods for VRP. We hope our work serves as a starting point for further advances in state-of-the-art models.

\bibliography{main}

\end{document}